\pdfoutput=1
\documentclass[11pt]{article}

\usepackage[final]{acl}

\usepackage[utf8]{inputenc}
\usepackage{times}
\usepackage{latexsym}
\usepackage{soul}
\usepackage{url}
\usepackage{graphicx}
\usepackage{amsmath}
\usepackage{amsthm}
\usepackage{booktabs}
\usepackage{algorithm}
\usepackage{algorithmic}
\usepackage[switch]{lineno}
\usepackage{subcaption}
\usepackage{caption}
\usepackage{color}
\usepackage{subfiles}
\usepackage{hyperref} % 
\usepackage{lipsum}
\usepackage{tcolorbox}
\tcbuselibrary{breakable}

\newtcolorbox{appendixbox}[1][]{
    colback=gray!10!white,  % 
    colframe=gray!50!black, % 
    arc=4pt,               % 
    boxrule=1pt,           % 
    left=6pt, right=6pt,   % 
    top=6pt, bottom=6pt,   % 
    breakable,             % 
    title={Appendix Prompt}, % 
    #1 % 
}
\usepackage[T1]{fontenc}
\usepackage[utf8]{inputenc}

\usepackage{microtype}

\usepackage{inconsolata}

\usepackage{graphicx}

\title{KnowPath: Knowledge-enhanced Reasoning via LLM-generated Inference Paths over Knowledge Graphs}

\author{\textbf{Qi Zhao}\textsuperscript{1}, 
\textbf{Hongyu Yang}\textsuperscript{1}, 
\textbf{Qi Song}\textsuperscript{1}\thanks{Corresponding author}, 
\textbf{Xinwei Yao}\textsuperscript{2}, 
\textbf{Xiangyang Li}\textsuperscript{1} \\
        \textsuperscript{1}University of Science and Technology of China, Hefei, Anhui, China \\ 
        \textsuperscript{2}Zhejiang University of Technology, Hangzhou, Zhejiang, China \\ 
        \texttt{\{zq2021, hongyuyang\}@mail.ustc.edu.cn} \\ 
        \texttt{xwyao@zjut.edu.cn, \{qisong09, xiangyangli\}@ustc.edu.cn}\\ 
}
\begin{document}
\maketitle

\begin{abstract}
    Large language models (LLMs) have demonstrated remarkable capabilities in various complex tasks, yet they still suffer from hallucinations. 
    % Introducing external knowledge, such as knowledge graph, can enhance the LLMs' ability to provide factual answers. 
    By incorporating external knowledge, such as knowledge graphs(KGs), LLMs can interactively explore KGs, thereby enhancing their ability to provide factual answers. This approach carries significant practical implications. 
    However, existing methods all suffer from three key limitations: insufficient mining of LLMs' internal knowledge, constrained generation of interpretable reasoning paths, and unclear fusion of internal and external knowledge. 
    Therefore, we propose KnowPath, a knowledge-enhanced large model framework driven by the collaboration of internal and external knowledge.
    It relies on the internal knowledge of the LLM to guide the exploration of interpretable directed subgraphs in external knowledge graphs, better integrating the two knowledge sources for more accurate reasoning. 
    Extensive experiments on multiple real-world datasets demonstrate the effectiveness of KnowPath.  
    Our code and data are available at: \url{https://github.com/tize-72/KnowPath}.
\end{abstract}
\section{Introduction}
\begin{figure}[t]
  \centering
    \includegraphics[width=\linewidth]{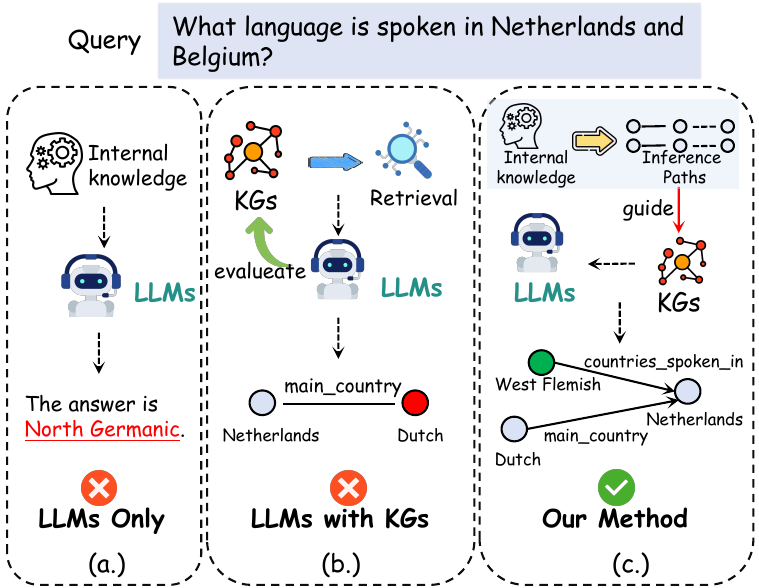}
  \caption{(a.) The LLMs-only approach suffers from severe hallucinations. (b.) The LLMs with KGs approach provides insufficient information, and their graph-based reasoning with KGs is often inaccurate. (c.) We first mine the internal knowledge of LLMs, offering more information for external KG reasoning and achieving better integration of internal and external knowledge in LLMs.}
  \label{fig:Insight}
\end{figure}
Large language models (LLMs) are increasingly being applied in various fields of Natural Language Processing (NLP) tasks, such as text generation~~\cite{text1, text2}, knowledge-based question answering~~\cite{chatkbqa, qa1}, and over specific domains~~\cite{special1, special2}. In most scenarios, LLMs serve as intermediary agents for implementing various functions~~\cite{agent1, agent2, agent3}. However, due to the characteristics of generative models, LLMs still suffer from hallucination issues, often generating incorrect answers that can lead to uncontrollable and severe consequences~~\cite{hallucinations1}. 
Introducing knowledge graphs (KGs) to mitigate this phenomenon is promising~\cite{survey}. This is because knowledge graphs store a large amount of structured factual knowledge, which can provide large models with accurate knowledge dependencies. At the same time, correcting the knowledge in large models often requires fine-tuning their model parameters, which inevitably incurs high computational costs~\cite{tog}. In contrast, updating knowledge graphs is relatively simple and incurs minimal overhead.

The paradigms of combining LLMs with KGs can be classified into three main categories. The first one is knowledge injection during pre-training or fine-tuning~\cite{rog, re-kbqa,unikgqa,givefact}. While the model's ability to grasp knowledge improves, these methods introduce high computational costs and catastrophic forgetting. 
The second one entails using LLMs as agents to reason through knowledge retrieved from the KGs. This approach does not require fine-tuning or retraining, significantly reducing overhead~\cite{structgpt, rag1}. 
However, its performance depends heavily on retrieval quality, where insufficient retrieval fails reasoning and excessive retrieval brings noise. 
The third one enables LLMs to participate in the process of knowledge exploration within external KGs~\cite{tog2}. In this case, the LLMs can engage in the selection of knowledge nodes at each step~\cite{tog, pog, gog}, thereby leveraging the advantages of the internal knowledge of the LLMs to some extent.

The effective patterns of LLMs introducing KGs still have limitations.
1) Insufficient exploration of internal knowledge in LLMs.
When exploring KGs, most approaches primarily treat LLMs as agents to select relevant relationships and entities, overlooking the potential of the internal knowledge. 
2) Constrained generation of interpretable reasoning paths. Some methods attempt to generate highly interpretable reasoning paths, but they limit the scale of path exploration, require additional memory. The generated paths also lack intuitive visual interpretability. 
3) Ambiguous fusion of internal and external knowledge. How to better integrate the internal knowledge of LLMs with the external knowledge in KGs still requires further exploration.

To overcome the above limitations, we propose KnowPath, a knowledge-enhanced large model framework driven by the collaboration of internal and external knowledge. 
Specifically, KnowPath consists of three stages.
1) Inference paths generation. To entirely exploit the internal knowledge of LLMs and adapt in zero-shot scenario, this stage employs a prompt-driven approach to extract the knowledge triples most relevant to the topic entities, and then generates reasoning paths based on these knowledge triples to attempt answering the question. 
2) Interpretable directed subgraph exploration. It refers to the process where the LLM combines the previously generated knowledge reasoning paths to select entities and relationships, and then responses based on the subgraph formed by these selections. This stage enables the LLMs to fully participate in the effective construction of external knowledge, while providing a clear process for constructing subgraphs.
3) Evaluation-based answering. At this stage, external knowledge primarily guides the KnowPath, while internal knowledge assists in generating the answer.
Our contributions can be summarized as follows:

\begin{itemize}
    \item We focus on a new view, emphasizing the importance of the LLMs' powerful internal knowledge in knowledge question answering, via a prompt-based internal knowledge reasoning path generation method for LLMs.
    \item We build a knowledge-enhanced LLM framework driven by the collaboration of internal and external knowledge. It not only integrates both the internal and external knowledge of the LLMs better, but also provides clearer and more interpretable reasoning paths.
    \item Experimental results on diverse knowledge QA datasets confirm that our proposed KnowPath delivers two key advantages: a marked reduction in LLM hallucinations and consistently better performance than existing state-of-the-art methods.
\end{itemize}

\section{KnowPath}
\subsection{Preliminary}
\textbf{Topic Entities} represent the main entities in a query $Q$, denoted as $e_0$. 
Each $Q$ contains $N$ topic entities $\{e_0^1, ...,  e_0^N\}$.

\begin{figure*}[t]
  \centering
    \includegraphics[width=0.98\linewidth]{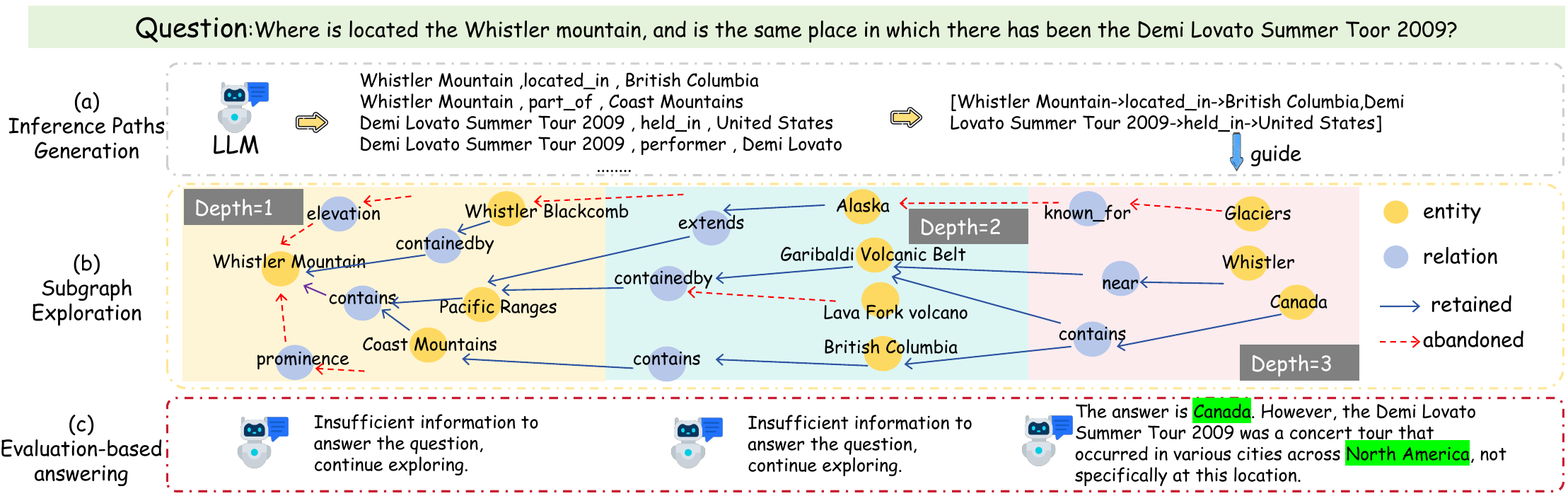}
  % \hspace{0.01\linewidth}
  \caption{The workflow of KnowPath. It contains: (a) Inference Paths Generation to exploit the internal knowledge of LLMs, (b)
  Subgraph Exploration to generate a interpretable directed subgraph, (c) Evaluation-based Answering to integrate internal and external knowledge.}
  \label{fig:workflow}
\end{figure*}

\noindent\textbf{Inference Paths} are a set of paths $P = {p_1, ..., p_L}$ generated by the LLM's own knowledge, where $L \in [1, N]$ is dynamically determined by the LLM agent. Each path $p$ starts from the topic entity $e_0 \in \{e_0^1,..., e_0^N\}$ and can be represented as $p = e_0 \to r_1 \to e_1 \to ... \to r_n \to e_n$, where $e_i$ and $r_i$ represent entities and relationships, respectively.

\noindent\textbf{Knowledge Graph (KG)} consists of many structured knowledge triples: $K=\{(e_h,r,e_t),r\in R,e_h,e_t\in E\}$, where $E$ represents all entities in the knowledge graph, and $R$ represents all relationships, and 
$e_h$ and $e_t$ represent the head and tail entities, respectively.

\noindent\textbf{KG Subgraph} refers to a connected subgraph extracted from the knowledge graph $K$, where the entities and relationships are entirely derived from $K$, i.e., $K_s\subseteq K$.

\subsection{Inference Paths Generation}
\label{sec:ipg}

Due to the extensive world knowledge stored within its parameters, LLMs can be considered as a complementary representation of KGs~\cite{llm-as-kg1, llm-as-kg2}.
To fully excavate the internal knowledge of LLMs and guide the exploration of KGs, we propose a prompt-driven method to extract the internal knowledge of LLMs effectively.
It can retrieve reasoning paths of the model's internal knowledge and clearly display the reasoning process, and also is particularly effective in zero-shot scenarios.
Specifically, given a query $Q$, we first guide the LLM to extract the most relevant topic entities $\{e_0^1, ..., e_0^N\}$ through a specially designed prompt.
Then, based on these topic entities, the large model is instructed to generate a set of knowledge triples associated with them. The number of triples $n$ is variable.
Finally, the LLM attempts to answer based on the previously generated knowledge triples and provides a specific reasoning path from entities and relations to the answer. Each path is in the form of $P=e_0^1 \to r_1 \to e_1 \to ... \to r_n \to e_n$. The details of the Inference Paths Generation process are presented in the Appendix \ref{prompt-inference}.

\subsection{Subgraph Exploration}

\textbf{Exploration Initialization.} 
KnowPath performs subgraph exploration for a maximum of $D$ rounds.
Each round corresponds to an additional hop in knowledge graph $K$ and the $j$-th contains $N$ subgraphs $\{K_{s,j}^1,...,K_{s,j}^N \}$.
Each subgraph $K_{s,j}^i$ is composed of a set of knowledge graph reasoning paths, i.e. $K_{s,j}^i=\{p_{1,j}^i\cup \cdots \cup p_{l,j}^i,i \in [1,N]\}$.
The number of reasoning paths $l$ is flexibly determined by the LLM agent.
Taking the $D$-th round and the $z$-th path as an example, it starts exploration from one topic entity $e_{0}^i$ and ultimately forms a connected subgraph of the KG, denoted as ${p_{z,D}^i=\{e_{0}^i, e_{1,z}^i,r_{1,z}^i, e_{2,z}^i, r_{2,z}^i, ..., r_{D,z}^i, e_{D,z}^i\}}$.
The start of the first round of subgraph exploration ($D$=0), each path $p_i$ corresponds to the current topic entity, i.e.${p_{z,0}^0=\{e_{0}^1\}}$.
% Similar to other works in the field, the topic entities in $Q$ are already aligned with the dataset.

\textbf{Relation Exploration.}
Relation exploration aims to expand the subgraphs obtained in each round of exploration, enabling deep reasoning. Specifically, for the $i$-th subgraph and the $j$-th round of subgraph exploration, the candidate entities are denoted as $E_{j}^{i}=\{e_{j-1,1}^i,...,e_{j-1,l}^i\}$, where $e_{j-1,1}^i$ is the tail entity of the reasoning path $p_{1,j-1}^i$.
Based on these candidates $E_{j}^{i}$, we search for all coresponding single-hop relations in knowledge graph $K$, denoted as $R_{a,j}^{i}=\{r_1,...,r_M\}$, where $M$ is determined by the specific knowledge graph $K$.
Finally, the LLM will rely on the query $Q$, the inference path $P$ generated through the LLM’s internal knowledge (Section \ref{sec:ipg}), and all topic entities $e_{0}$ to select the most relevant candidate relations from $R_{a,j}^{i}$, denoted as $R_{j}^{i} \subseteq R_{a,j}^{i} $, which is dynamically determined by the LLM agent. The prompt is shown in Appendix \ref{prompt-Relation}.

\textbf{Entity Exploration.}
Entity exploration depends on the already determined candidate entities and candidate relations.
Taking the $i$-th subgraph and the $j$-th round of subgraph exploration as an example, relying on $E_{j}^{i}$ and $R_{j}^{i}$, we perform queries like $(e, r, ?)$ or $(?, r, e)$ on the knowledge graph $K$ to retrieve the corresponding entities $E_{a,j}^{i} = \{e_1, ..., e_N\}$, where $N$ varies depending on the knowledge graph $K$.
Then, the agent also considers the query $Q$, the inference path $P$ in Section \ref{sec:ipg}, the topic entity $e_{0}^i$, and the candidate relation set $R_{j}^{i}$ from $E_{a,j}^{i}$ to generate the most relevant entity set $E_{j+1}^{i} = \{e_{j,1}^i, ..., e_{j,l}^i\} \subseteq E_{a,j}^{i}$. Note that $e_{j,1}^i$ is the tail entity of the reasoning path $p_{1,j}^i$. The prompt is shown in Appendix \ref{prompt-Entity}.

\textbf{Subgraph Update.}
Relation exploration determines entity exploration, and we update the subgraph only after completing the entity exploration. The subgraph exploration algorithm can be found in Algorithm~\ref{alg:update_Subgraph} in the Appendix \ref{appendix-subgraph-update}. 
Specifically, for the $i$-th subgraph and the $j$-th round of subgraph exploration, we append the result of the exploration $( , r, e_{j,1}^i)$ to the path $p_{1,j}^i$ in the subgraph $K_{s,j}^i$.
This path update algorithm not only considers the directionality of entities and relations, but also automatically determines and updates the paths.
The detailed process is described in Algorithm~\ref{alg:update_path} in the Appendix \ref{appendix-Path-update}.
The final subgraph can be flexibly expanded due to the variable number of paths $l$.

\subsection{Evaluation-based Answering}

After completing the subgraph update for each round, the LLM attempts to answer the query through the subgraph $\{K_{s,j}^1, ..., K_{s,j}^N \}$.
If it determines that the current subgraph is insufficient to answer the question, the next round of subgraph exploration will be executed, until the maximum exploration depth $D$ is reached.
Otherwise, it will output the final answer along with the corresponding interpretable directed subgraph.
Unlike previous work ~\cite{pog}, even if no answer is found at the maximum exploration depth, KnowPath will rely on the inference path $P$ to response.
The framework of KnowPath is shown in Figure~\ref{fig:workflow}. The prompt is shown in Appendix \ref{prompt-Evaluation}.
% generated through the LLM’s internal knowledge (Section \ref{sec:ipg}) to answer the query.

\section{Experimental Setup}\label{sec:exp}
\begin{table*}[ht]
\centering
\tabcolsep=0.35cm
\resizebox{\textwidth}{!}{
\begin{tabular}{lcccc}
\toprule
\textbf{Method}              & \textbf{CWQ} & \textbf{WebQSP} & \textbf{Simple Questions} & \textbf{WebQuestions} \\ 
\midrule
\multicolumn{5}{c}{\textbf{LLM only}} \\
\midrule
IO prompt~~\cite{ioprompt}                    & 37.6 $\pm$ 0.8         & 63.3 $\pm$ 1.2            & 20.0 $\pm$ 0.5              & 48.7 $\pm$ 1.4               \\ 
COT~~\cite{cot}                          & 38.8 $\pm$ 1.5         & 62.2 $\pm$ 0.7            & 20.5 $\pm$ 0.4              & 49.1 $\pm$ 0.9               \\ 
RoG w/o planning~~\cite{rog}             & 43.0 $\pm$ 0.9         & \cellcolor{blue!19}66.9 $\pm$ 1.3            & -                 & -                  \\ 
SC~~\cite{sc}                           & \cellcolor{blue!19}45.4 $\pm$ 1.1         & 61.1 $\pm$ 0.5            & 18.9 $\pm$ 0.6              & \cellcolor{blue!19}50.3 $\pm$ 1.2               \\ 
\midrule
\multicolumn{5}{c}{\textbf{Fine-Tuned KG Enhanced LLM}} \\ 
\midrule
UniKGQA~~\cite{unikgqa}           & 51.2 $\pm$ 1.0         & 75.1 $\pm$ 0.8          & -        & -         \\
RE-KBQA~~\cite{re-kbqa}  & 50.3 $\pm$ 1.2         & 74.6 $\pm$ 1.0            & -                 & -        \\
ChatKBQA~~\cite{chatkbqa}                     & 76.5 $\pm$ 1.3         & 78.1 $\pm$ 1.1            & \cellcolor{blue!19}85.8 $\pm$ 0.9              & 55.1 $\pm$ 0.6                  \\ 
RoG~~\cite{rog}     & \cellcolor{blue!19}64.5 $\pm$ 0.7         & \cellcolor{blue!19}85.7 $\pm$ 1.4            & 73.3 $\pm$ 0.8    & \cellcolor{blue!19}56.3 $\pm$ 1.0               \\ 
\midrule
\multicolumn{5}{c}{\textbf{Prompting KG Enhanced LLM with GPT3.5}} \\ 
\midrule
StructGPT~~\cite{structgpt}      & 54.3 $\pm$ 1.0         & 72.6 $\pm$ 1.2        & 50.2 $\pm$ 0.5      & 51.3 $\pm$ 0.9          \\
ToG~~\cite{tog}                          & 57.1 $\pm$ 1.5         & 76.2 $\pm$ 0.8            & 53.6 $\pm$ 1.0              & 54.5 $\pm$ 0.7               \\ 
PoG~~\cite{pog}                          & 63.2 $\pm$ 1.0         & 82.0 $\pm$ 0.9            & 58.3 $\pm$ 0.6              & 57.8 $\pm$ 1.2               \\ 
\textbf{KnowPath (Ours)}                   & \cellcolor{blue!19}\textbf{67.9 $\pm$ 0.6} & \textbf{84.1 $\pm$ 1.3}   & \cellcolor{blue!19}\textbf{61.5 $\pm$ 0.8}     & \cellcolor{blue!19}\textbf{60.0 $\pm$ 1.0}      \\
\midrule
\multicolumn{5}{c}{\textbf{Prompting KG Enhanced LLM with DeepSeek-V3}} \\ 
\midrule
ToG~~\cite{tog}           & 60.9 $\pm$ 0.7     & 82.6 $\pm$ 1.0         & 59.7 $\pm$ 0.9      & 57.9 $\pm$ 0.8                  \\ 
PoG~~\cite{pog}              & 68.3 $\pm$ 1.1        & 85.3 $\pm$ 0.9         & 63.9 $\pm$ 0.5       & 61.2 $\pm$ 1.3           \\ 
\textbf{KnowPath (Ours)}     & \cellcolor{blue!19}\textbf{73.5 $\pm$ 0.9}     & \cellcolor{blue!19}\textbf{89.0 $\pm$ 0.8}        & \cellcolor{blue!19}\textbf{65.3 $\pm$ 1.0}       & \cellcolor{blue!19}\textbf{64.0 $\pm$ 0.7}       \\ 
\bottomrule
\end{tabular}
}
\caption{Hits@1 scores (\%) of different models on four datasets under various knowledge-enhanced methods. We use GPT-3.5 and DeepSeek-V3 as the primary backbones. \textbf{Bold text} indicates the results achieved by our method.}
\label{tab:comparison}
\end{table*}

\subsection{Baselines}

We chose corresponding advanced baselines for comparison based on the three main paradigms of existing knowledge-based question answering.
1) The First is the LLM-only, including the standard prompt (IO prompt~\cite{ioprompt}), the chain of thought prompt (CoT~\cite{cot}), the self-consistency (SC~\cite{sc}), and the RoG without planning (ROG w/o planning~\cite{rog}).
2) The second is the KG-enhanced fine-tuned LLMs, which include ChatKBQA~\cite{chatkbqa}, RoG~\cite{rog}, UniKGQA~\cite{unikgqa}, and RE-KBQA~\cite{re-kbqa}.
3) The third is the KG-enhanced prompt-based LLMs, including Think on graph (ToG~\cite{tog}), Plan on graph (PoG~\cite{pog}), and StructGPT~\cite{structgpt}. 
Unlike the second, this scheme no longer requires fine-tuning and has become a widely researched mode today.

\subsection{Datasets and Metrics}
\textbf{Datasets.}
We adopt four knowledge-based question answering datasets: the single-hop Simple Questions~~\cite{simpleqa}, the complex multi-hop CWQ~~\cite{cwq} and WebQSP~~\cite{webqsp}, and the open-domain WebQuestions~~\cite{webquestion}.  Detailed descriptions are provided in Appendix \ref{dataset}.

\noindent\textbf{Metrics.}
Following previous research ~~\cite{pog}, we apply exact match accuracy (Hits@1) for evaluation.

\subsection{Experiment Details}
Following previous research ~~\cite{pog}, to control the overall costs, the maximum subgraph exploration depth $D_{amx}$ is set to 3. Since the FreeBase~~\cite{freebase} supports all the aforementioned datasets, we apply it as the base graph for subgraph exploration, and We apply GPT-3.5-turbo-1106 and DeepSeek-V3 as the base models.
All experiments are deployed on four NVIDIA A800-40G GPUs. The prompts and SPARQL queries used in the experiments can be found in Appendices \ref{appendix-prompts} and \ref{Sparql}, respectively.
\section{Result}

\subsection{Main results}

We conducted comprehensive experiments on four widely used knowledge-based question answering datasets. The experimental results are presented in Table \ref{tab:comparison}, and four key findings are outlined as follows:

\textbf{KnowPath achieves state-of-the-art results.}
Our KnowPath outperforms all the Prompting-driven KG-Enhanced.
For instance, on the multi-hop CWQ, regardless of the base model used, KnowPath achieves a maximum improvement of about 13\% in Hits@1.
In addition, KnowPath outperforms the LLM-only with a clear margin and surpasses the majority of Fine-Tuned KG-Enhanced LLM methods.
On the most challenging open-domain question answering dataset WebQuestions, KnowPath achieves the best performance compared to strong baselines from other paradigms (e.g., PoG 61.2\% vs Ours 64.0\%). This demonstrates KnowPath's ability to enhance the factuality of LLMs in open-domain question answering, which is an intriguing phenomenon worth further exploration.

\textbf{KnowPath excels at complex multi-hop tasks.}
On both CWQ and WebQSP, KnowPath outperforms the latest strong baseline PoG, achieving an average improvement of approximately 5\% and 2.9\%, respectively.
On the WebQSP, DeepSeek-v3 with KnowPath not only outperforms all Prompting-based KG-Enhanced LLMs but also surpasses the strongest baseline ROG among Fine-Tuned KG-Enhanced LLMs (85.7\% vs 89\%). 
On the more challenging multi-hop CWQ, the improvement of KnowPath over the PoG is significantly greater than the improvement on the simpler single-hop SimpleQuestions (5.2\% vs 1.4\%).
% These collectively indicate that KnowPath is sensitive to deep reasoning.

\textbf{Knowledge enhancement greatly aids factual question answering.}
When question answering is based solely on LLMs, the performance is poor across multiple tasks. For example, COT achieves only about 20.5\% Hits@1 on SimpleQuestions.
This is caused by the hallucinations inherent in LLMs.
Whatever method is applied to introduce the KGs, they significantly outperform LLM-only. 
The maximum improvements across the four tasks are 35.9\%, 27.9\%, 46.4\%, and 15.3\%. 
% with the least improvements being 22.5\%, 17.2\%, 41\%, and 9.7\%, respectively.
These further emphasize the importance of introducing knowledge graphs for generating correct answers.

\textbf{The stronger the base, the higer the performance.}
As DeepSeek-V3 is better than GPT-3.5, even though both are prompting-based knowledge-enhanced, their performance on all tasks shows a significant difference after incorporating our KnowPath.
Replacing GPT-3.5 with DeepSeek-V3, KnowPath achieved a maximum improvement from 67.9\% to 73.5\% on CWQ, and on Simple Questions, it improved by at least 3.8\%.
These findings indicate that the improvement in model performance directly drives the enhancement of its performance in knowledge-based question-answering.

\textbf{KnowPath is a more flexible plugin.}
Compared to fine-tuned knowledge-enhanced LLMs, our KnowPath does not require fine-tuning of the LLM, yet it outperforms most of the fine-tuned methods.
In addition, on the CWQ dataset, KnowPath with DeepSeek-V3 achieves performance that is very close to the strongest baseline, ChatKBQA, which requires fine-tuning for knowledge enhancement. On the WebQSP dataset, it outperforms ChatKBQA by about 11\% (78.1\% vs 89.0\%).
Overall, the resource consumption of KnowPath is significantly lower than that of Fine-Tuned KG-Enhanced LLMs.
This is because KnowPath improves performance by optimizing inference paths and enhancing knowledge integration, making it a more flexible and plug-and-play framework.

% 消融实验表格
\begin{table}[htbt]
\centering
\resizebox{\columnwidth}{!}{  % 将表格宽度调整为单栏宽度
\begin{tabular}{lcccc}
\toprule
\textbf{Method} & \textbf{CWQ} & \textbf{WebQSP} & \textbf{SimpleQA} & \textbf{WebQ} \\
\midrule
KnowPath & 73.5 & 89.0 & 65.3 & 64.0 \\
-w/o IPG & 67.3 & 84.5 & 63.1 & 61.0 \\
-w/o SE & 64.7 & 83.1 & 60.4 & 60.7 \\
Base & 39.2 & 66.7 & 23.0 & 53.7 \\
\bottomrule
\end{tabular}
}
\caption{Ablation experiment results on four knowledge-based question answering tasks. IPG stands for Inference Paths Generation module, while SE stands for Subgraph Exploration module.}
\label{table:ablation}
\end{table}
\begin{table}[htb]
\centering
\resizebox{\columnwidth}{!}{  % 
\begin{tabular}{lccccc}
\toprule
\textbf{Method} & \textbf{LLM Call} & \textbf{Total Token} & \textbf{Input Token} &\textbf{Time(s)}\\
\midrule
ToG & 22.6 & 9669.4 & 8182.9 & 96.5\\
PoG & 16.3 & 8156.2 & 7803.0 & 23.3\\
KnowPath & \textbf{9.9} & \textbf{2742.4} & \textbf{2368.9} & \textbf{16.5} \\
\bottomrule
\end{tabular}
}
\caption{Cost-effectiveness analysis on the CWQ dataset between our KnowPath and the strongly prompt-driven knowledge-enhanced benchmarks (ToG and PoG). The Total Token includes two parts: the total number of tokens from multiple input prompts and the total number of tokens from the intermediate results returned by the LLM. The Input Token represents only the total number of tokens from the multiple input prompts. The LLM Call refer to the total number of accesses to the LLM agent, and Time represents the reasoning time. }
\label{table:tokens}
\end{table}
\begin{figure}[ht]
  \centering
  \includegraphics[width=0.94\linewidth]{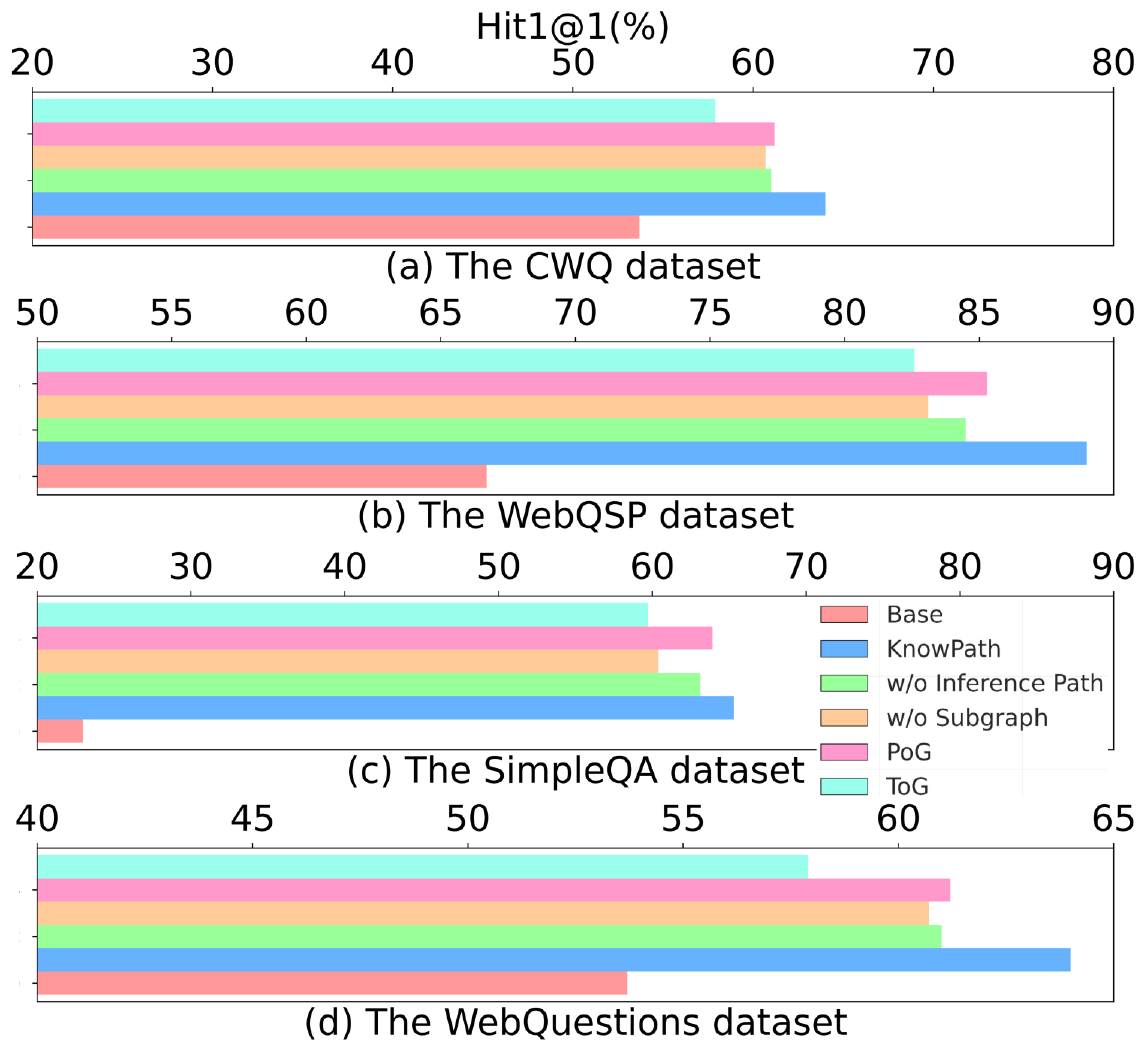}
  \caption{Comparison of KnowPath, its individual components, and strong baseline methods (ToG and PoG) on the performance across four commonly used knowledge-based question answering datasets.}
  \label{fig:ablation}
\end{figure}
\subsection{Ablation Study}
We validate the effectiveness of each component of KnowPath and quantify their contributions to performance.
Its results are presented in Table \ref{table:ablation}, and visualized in Figure \ref{fig:ablation}.
\begin{figure*}[ht]
  \centering
  \includegraphics[width=0.24\linewidth]{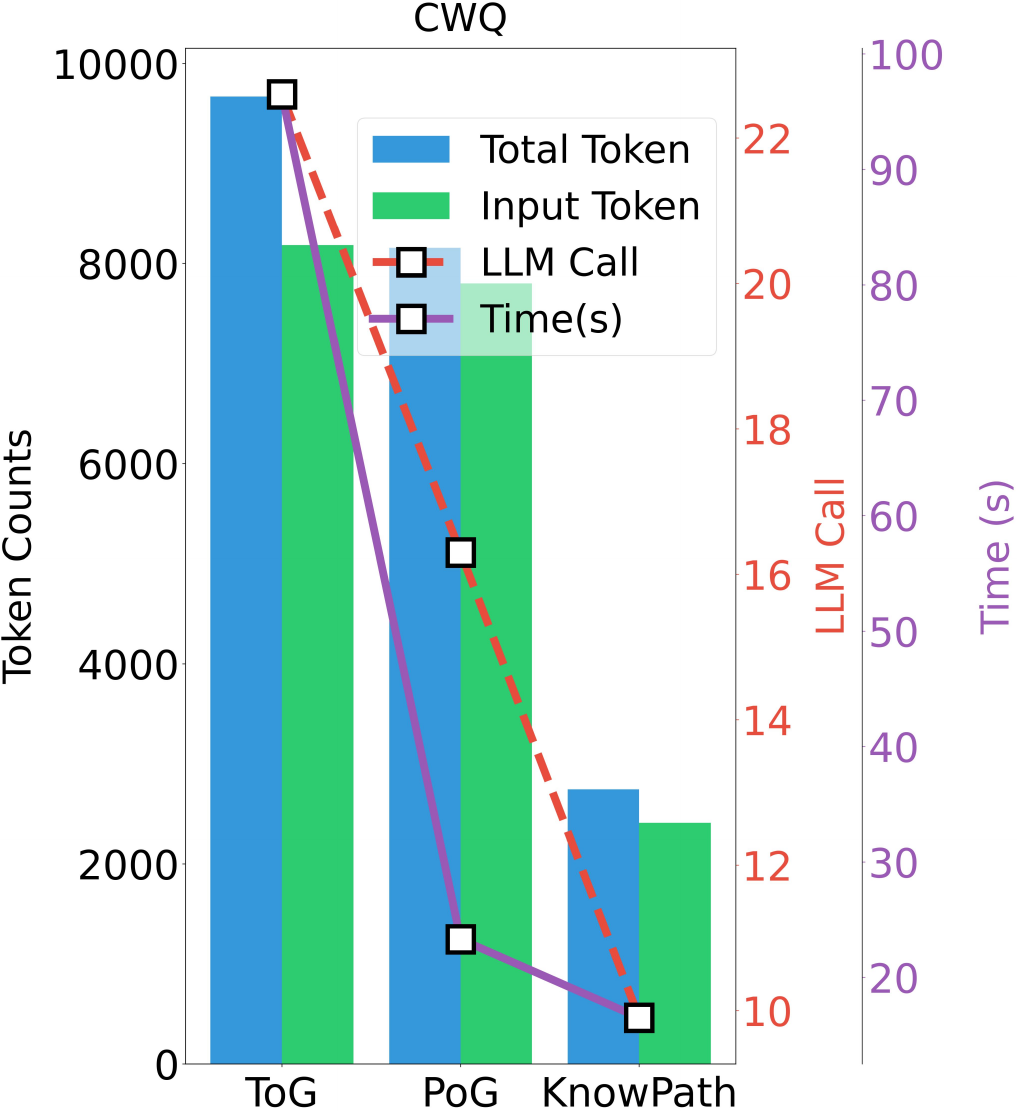}
  % \hspace{0.01\linewidth}
  \includegraphics[width=0.24\linewidth]{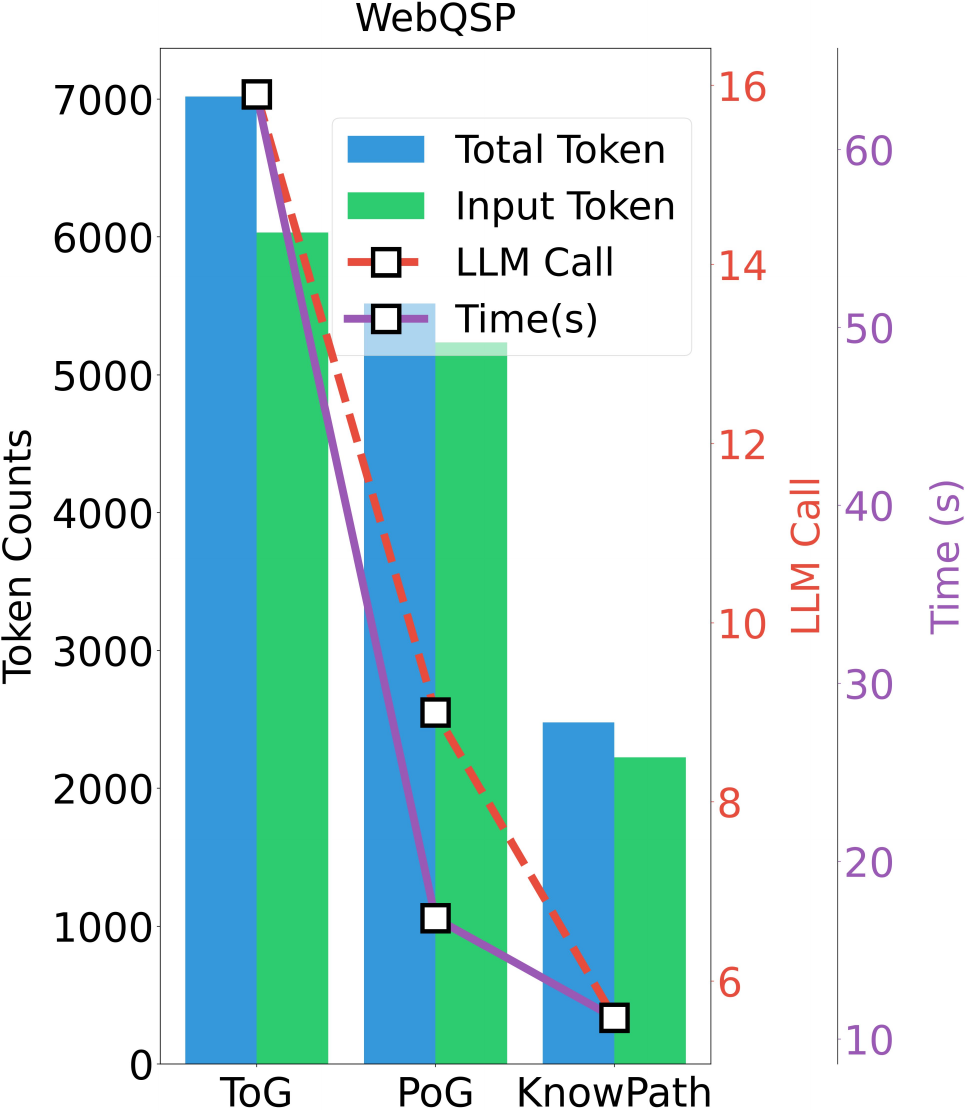}
  \hspace{0.01\linewidth}
  \includegraphics[width=0.24\linewidth]{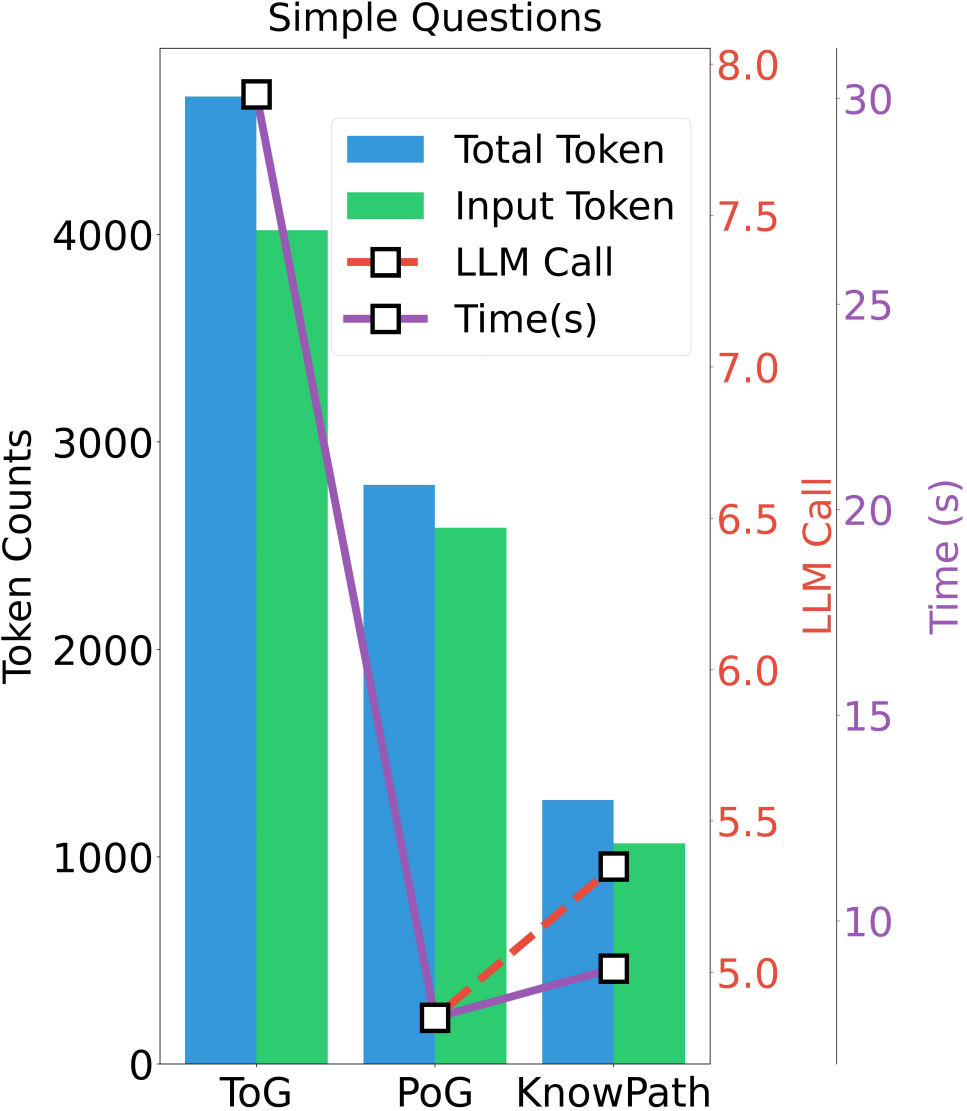}
  % \hspace{0.01\linewidth}
  \includegraphics[width=0.24\linewidth]{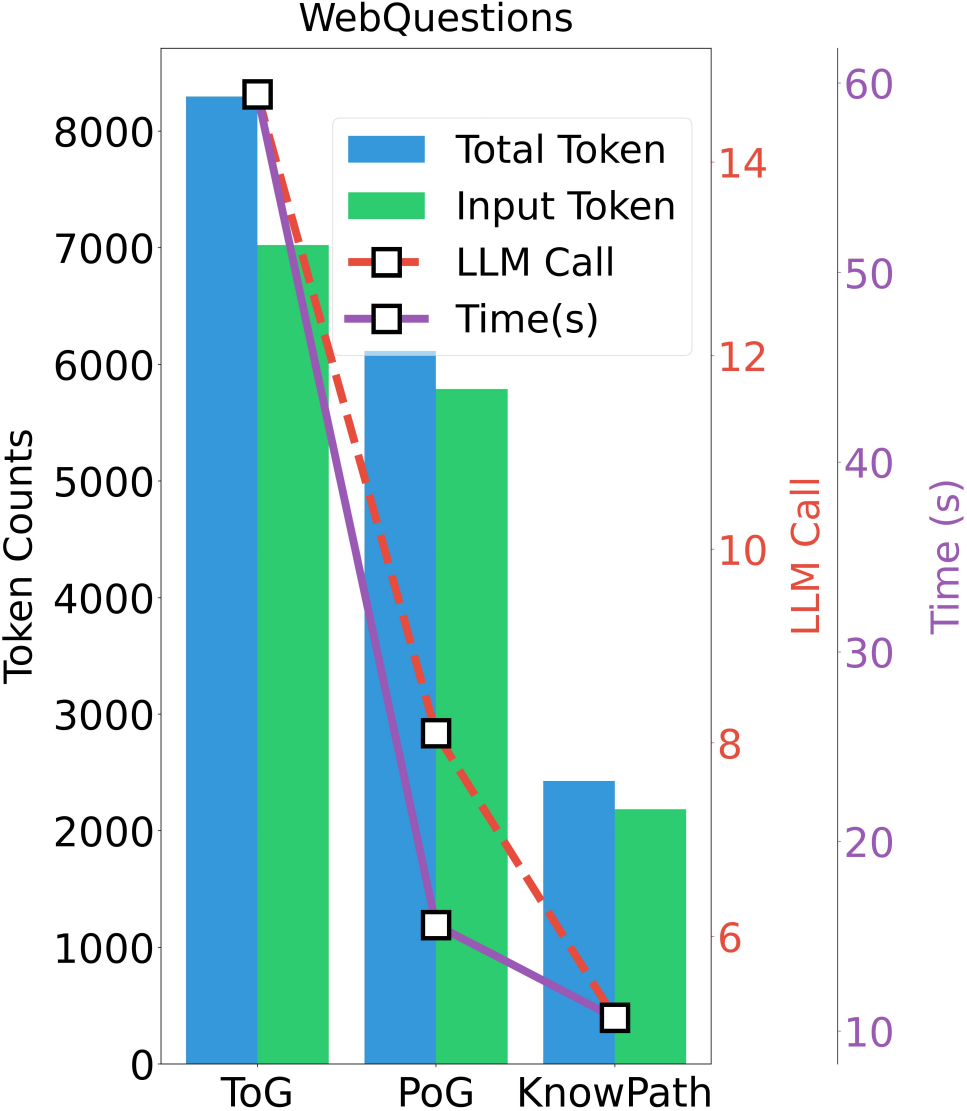}
  \caption{Visualization of the cost-effectiveness analysis on four knowledge-based question-answering datasets.}
  \label{fig:efficiency-analysis}
\end{figure*}
\begin{figure}[t]
  \centering
  \begin{subfigure}{0.49\linewidth}
    \includegraphics[width=\linewidth]{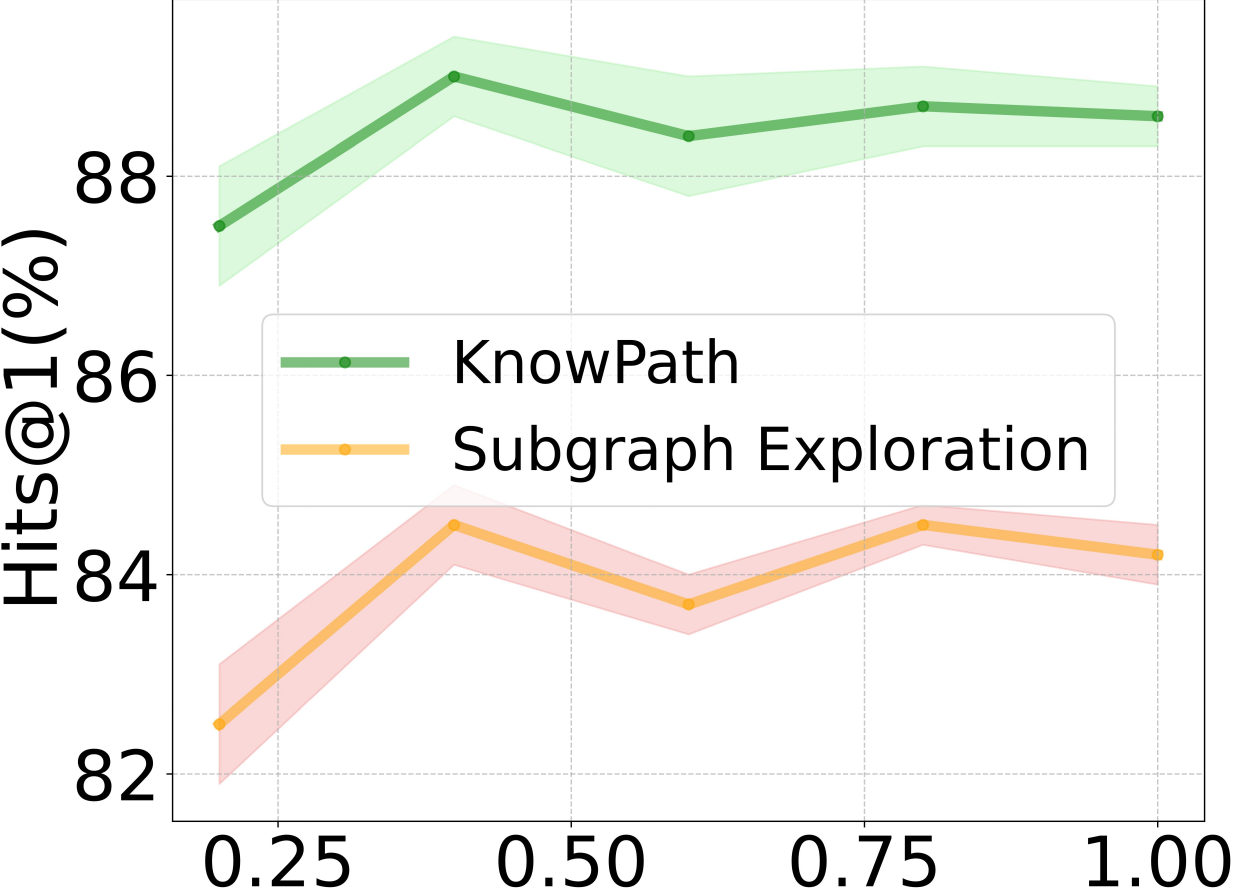}
    \caption{Exploration temperature}
      \label{fig:tempreture}
  \end{subfigure}
  \begin{subfigure}{0.49\linewidth}
    \includegraphics[width=\linewidth]{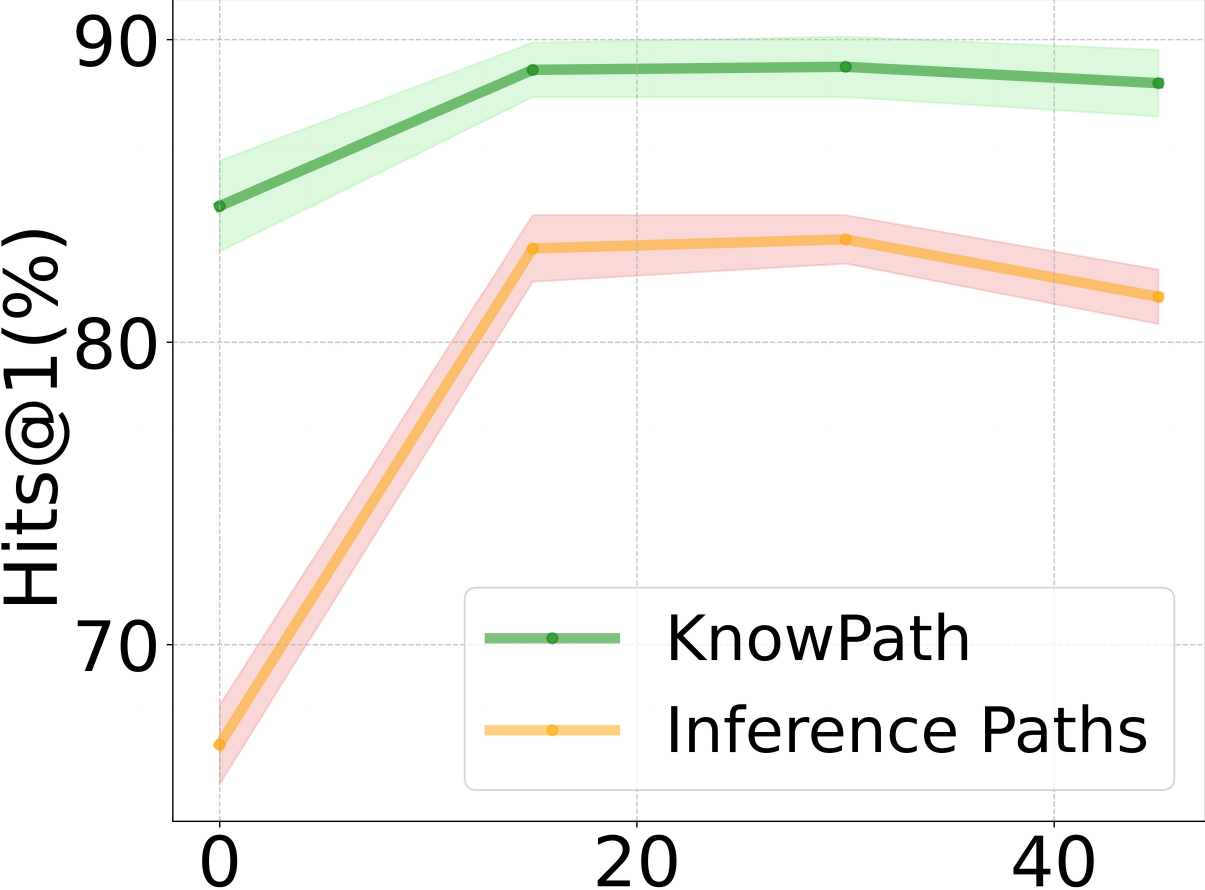}
    \caption{The count of triples}

    \label{fig:triple}
  \end{subfigure}
  % \hspace{0.01\linewidth}
  \caption{Analysis of key parameters.}
  \label{fig:Parameter}
\end{figure}

\begin{figure*}[t]
  \centering
    \includegraphics[width=0.98\linewidth]{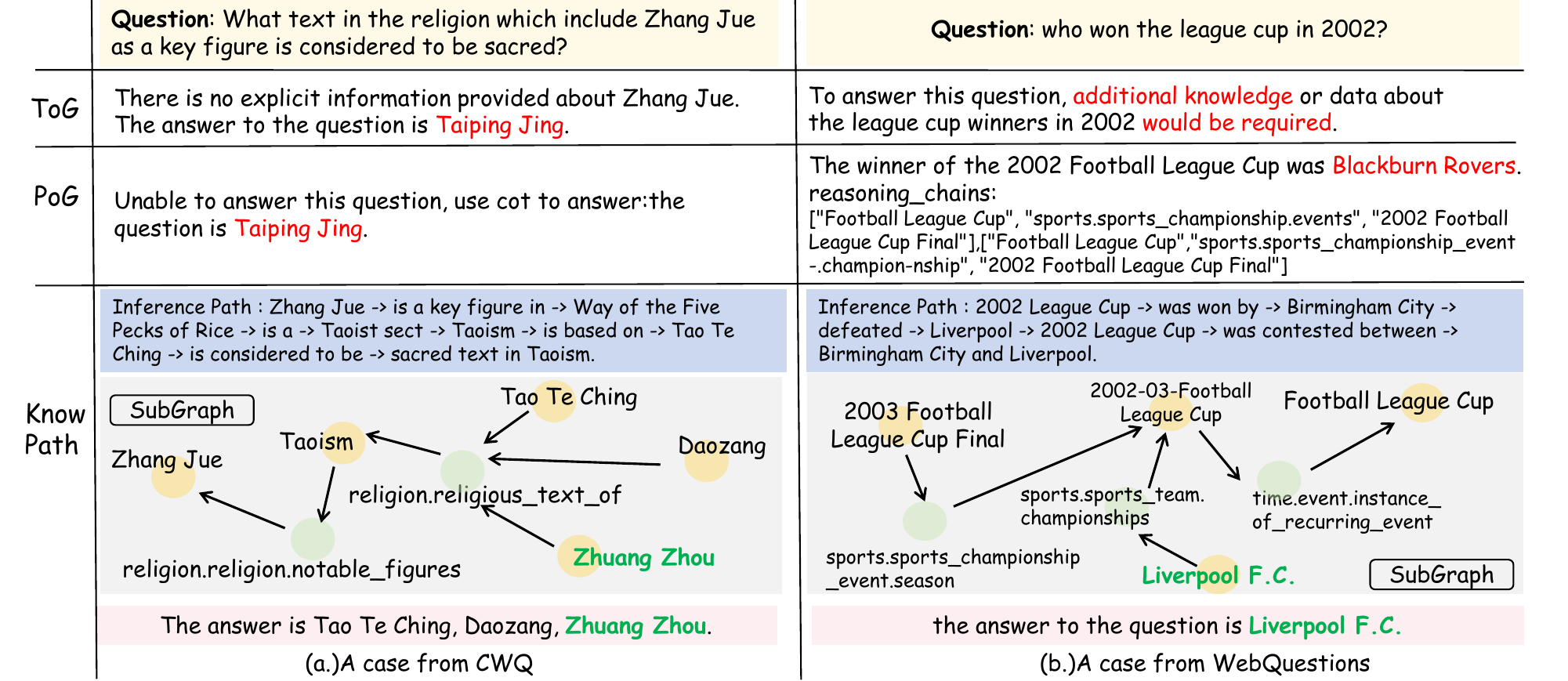}
  % \hspace{0.01\linewidth}
  \caption{The case study on the multi-hop CWQ and open-domain WebQuestions dataset. To provide a clear and vivid comparison with the strong baselines (ToG and PoG), we visualized the execution process of KnowPath}
  \label{fig:case}
\end{figure*}

\textbf{Each component contributes to the overall remarkable performance.}
After removing each module, their performance on different datasets will decline. However, compared to the base model, the addition of these modules still significantly improves the overall performance.

\textbf{It is necessary to focus on the powerful internal knowledge of LLMs.}
Eliminating the Subgraph Exploration and relying solely on the internal knowledge mining of LLMs to generate reasoning paths and provide answers proves to be highly effective. 
It has shown significant improvement across all four datasets, with an average performance enhancement of approximately 21.6\%. The most notable improvement was observed on SimpleQA, where performance leaped from 23\% to 60.4\%.
This indicates that even without the incorporation of external knowledge graphs, the performance of the model in generating factual responses can be enhanced to a certain extent through internal mining methods.
However, without the guidance of internal knowledge reasoning paths, KnowPath has seen some performance decline across all tasks, especially in complex multi-hop CWQ and WebQSP.

\textbf{The most critical credible directed Subgraph Exploration is deep-sensitive.}
Removing the subgraph exploration leads to a significant decline in Knowpath across all tasks, averaging a drop of approximately 5.7\%. This performance dip is particularly pronounced in complex multi-hop tasks. For instance, on the CWQ, Knowpath without subgraph exploration experiences a nearly 9\% decrease.

\subsection{Cost-effectiveness Analysis}

To evaluate KnowPath’s cost-effectiveness while maintaining high accuracy, we conduct a comprehensive cost-benefit analysis. This experiment systematically tracks primary cost drivers, including LLM API calls, token consumption, and inference time. Quantitative results are tabulated in Table \ref{table:tokens}, with complementary visualizations presented in Figure \ref{fig:efficiency-analysis}. Our key findings are described as follows:

% \textbf{LLM agent calls was significantly reduced.} Specifically, the LLM calls for TOG and POG was 2.28x and 1.64x of that in our KnowPath, respectively.
% This exceptionally low cost can be attributed to the fact that the Subgraph Exploration does not limit the scale of the path search, and this can be broken down into three key reasons. First, in each round of subgraph exploration, only one relation exploration and one entity exploration are conducted. Second, the Evaluation-based answering only accesses the LLM once after each round of subgraph exploration to judge whether the current subgraph can answer the question. If it cannot, the next round is performed. Third, if the largest explored subgraph still cannot answer the question, KnowPath will rely on the Inference Paths Generation.
\textbf{KnowPath achieves significant reduction in LLM call while simultaneously decreasing inference latency.} Specifically, TOG and POG require 2.28× and 1.64× more LLM invocations on average compared to KnowPath. This efficiency improvement stems from three key design features: (1) During each subgraph exploration round, KnowPath performs only a single relation exploration followed by a single entity exploration; (2) The evaluation module queries the LLM only once per round to determine if the current subgraph can answer the question, proceeding to the next round only when necessary; (3) When the maximally expanded subgraph remains insufficient for question answering, , KnowPath will rely on the Inference Paths Generation to produce answers.

\textbf{The number of tokens used is saved by several times.}
Whether in Total Token or Input Tokens, KnowPath saves approximately 4.0x compared to TOG and POG. 
This is mainly since all the prompts used in KnowPath are based on the carefully designed zero-shot approach, rather than the in-context learning used by the previous, which require providing large context to ensure the factuality of the answers. 
% We explored the reasons behind this difference.
% First, previous methods rely on more contextual information for in-context learning to ensure the correctness of the output.
% Secondly, KnowPath fully leverages the powerful internal relevant knowledge and uses it as the input signal for the agent.
% This not only provides more contextual reference but also significantly improves the accuracy and efficiency of relation and entity exploration in subgraph exploration, ensuring that the generated subgraph is highly relevant while enabling the most effective reasoning toward potential answers.
We investigate the reasons for this discrepancy. KnowPath fully leverages the model's robust internal knowledge, utilizing it as input signals for the agent. This approach provides richer contextual references, significantly improving both accuracy and efficiency in relation/entity exploration during subgraph traversal. The resulting subgraphs maintain high relevance while enabling optimal reasoning about potential answers. The comprehensive performance evaluation of KnowPath across the additional three datasets is provided in Appendix \ref{Efficiency}.

\subsection{Parameter analysis}

We analyze the key parameters that affect the performance of KnowPath on the WebQSP, and discuss the following issues:

\textbf{What is the impact of the temperature in Subgraph Exploration?}
We explore the optimal temperature from 0.2 to 1, and the relation between it and Hits@1 is shown in Figure \ref{fig:tempreture}.
During subgraph exploration, variations in the temperature affect the divergence of the model's generated answers. A lower temperature negatively impacts KnowPath's performance, as the model generates overly conservative answers with insufficient knowledge, while the LLM relies on its internal knowledge when exploring and selecting entities and relationships. A higher temperature also harms KnowPath, as the divergent answers may deviate from the given candidates.
Extensive experiments show that 0.4 is the optimal temperature, consistent with other existing works~~\cite{pog}.

\textbf{How is the count of knowledge triples determined in Inference Paths Generation?}
We explored it with a step size of 15, and the relationship between the count of knowledge triples and Hits@1 is shown in Figure \ref{fig:tempreture}.
When the count is 0, KnowPath's performance is poor due to the lack of internal knowledge exploration. When the count is too large, such as 45, its performance is also suboptimal, as excessive exploration introduces irrelevant knowledge as interference. Extensive experiments show that 15 is the optimal.

\subsection{Case Study}

To provide a clear and vivid comparison with the strong baselines, we visualized the execution process of KnowPath, as shown in Figure \ref{fig:case}. 
In the CWQ, ToG and PoG can only extract context from the question, failing to gather enough accurate knowledge for a correct answer, thus producing the incorrect answer "Taiping Jing." In contrast, KnowPath uncovers large model reasoning paths that provide additional, sufficient information. This enables key nodes, such as "Taoism," to be identified during subgraph exploration, ultimately leading to the correct answer, "Zhuang Zhou."
In the WebQuestions, ToG is unable to answer the question due to insufficient information. 
Although PoG provides a reasoning chain, the knowledge derived from the reasoning process is inaccurate, and the final answer still relies on the reasoning of the large model, resulting in the incorrect answer "Blackburn Rovers."
In contrast, guided by Inference Path, KnowPath accurately identified the relationship "time.event.instance\_of\_recurring\_event" and, through reasoning with the node "2002-03-Football League Cup," ultimately arrived at the correct result node "Liverpool F.C."
Overall, KnowPath not only provides answers but also generates directed subgraphs, and significantly enhance the interpretability of the results. Another visualization of the generated subgraph can be found in Appendix \ref{appendix-Visualize}.
\section{Related Work}

\textbf{Prompt-driven LLM inference.} CoT~\cite{cot} (Chain of Thought) effectively improves the reasoning ability of large models, enhancing performance on complex tasks with minimal contextual prompts. Self-Consistency (SC)~\cite{sc} samples multiple reasoning paths to select the most consistent answer, with further improvements seen in DIVERSE~\cite{cot-more-1} and Vote Complex~\cite{cot-more-2}. Other methods have explored CoT enhancements in zero-shot scenarios~\cite{zero1,zero2}. However, reasoning solely based on the model's knowledge still faces significant hallucination issues, which remain unresolved.

\textbf{KG-enhanced LLM inference.} "Early works enhanced model knowledge understanding by injecting KGs into model parameters through fine-tuning or retraining~\cite{re-kbqa,unikgqa,givefact}. ChatKBQA~\cite{chatkbqa} and RoG~\cite{rog} utilize fine-tuned LLMs to generate logical forms. StructGPT~\cite{structgpt}, based on the RAG approach, retrieves information from KGs for question answering. ToG~\cite{tog} and PoG~\cite{pog} involve LLMs in knowledge graph reasoning, using them as agents to assist in selecting entities and relationships during exploration. Despite achieving strong performance, these methods still face challenges like insufficient internal knowledge mining and the inability to generate interpretable reasoning paths.

\section{Conclusion}

In this paper, we propose the knowledge-enhanced reasoning framework KnowPath, driven by the collaboration of internal and external knowledge.
It focuses on leveraging the reasoning paths generated by the extensive internal knowledge of LLMs to guide the interpretable directed subgraph exploration of knowledge graphs.
Extensive experiments show that:
1) Our KnowPath is optimal and excels at complex multi-hop tasks.
2) It demonstrates remarkable cost-effectiveness, with a 55\% reduction in the number of LLM calls and a 75\% decrease in the number of tokens consumed compared to the strong baselines.
3) KnowPath can explore directed subgraphs of the KGs, providing an intuitive and interpretable reasoning process, greatly enhancing the overall interpretability.

\section*{Limitations}
In this work, we propose KnowPath. We show that existing methods cannot effectively combine internal and external knowledge in LLMs, and we introduce reasoning paths generated from the model's own knowledge to improve its performance in exploring external knowledge graphs.

KnowPath still has some limitations. First, like existing related approaches, KnowPath requires multiple rounds of interaction with external knowledge graphs during question answering, which introduces time overhead. Reducing this latency remains a challenge for real-time QA scenarios. Second, we only tested knowledge QA with single text modality. How to incorporate images, audio, and videos for testing requires future research.

%

% \section*{Acknowledgments}

% This document has been adapted
% by Steven Bethard, Ryan Cotterell and Rui Yan
% from the instructions for earlier ACL and NAACL proceedings, including those for
% ACL 2019 by Douwe Kiela and Ivan Vuli\'{c},
% NAACL 2019 by Stephanie Lukin and Alla Roskovskaya,
% ACL 2018 by Shay Cohen, Kevin Gimpel, and Wei Lu,
% NAACL 2018 by Margaret Mitchell and Stephanie Lukin,
% Bib\TeX{} suggestions for (NA)ACL 2017/2018 from Jason Eisner,
% ACL 2017 by Dan Gildea and Min-Yen Kan,
% NAACL 2017 by Margaret Mitchell,
% ACL 2012 by Maggie Li and Michael White,
% ACL 2010 by Jing-Shin Chang and Philipp Koehn,
% ACL 2008 by Johanna D. Moore, Simone Teufel, James Allan, and Sadaoki Furui,
% ACL 2005 by Hwee Tou Ng and Kemal Oflazer,
% ACL 2002 by Eugene Charniak and Dekang Lin,
% and earlier ACL and EACL formats written by several people, including
% John Chen, Henry S. Thompson and Donald Walker.
% Additional elements were taken from the formatting instructions of the \emph{International Joint Conference on Artificial Intelligence} and the \emph{Conference on Computer Vision and Pattern Recognition}.

\bibliography{knowpath}

\appendix
\label{sec:appendix}
\section{Subgraph Exploration Algorithm}\label{appendix-subgraph-update}

Here we present the subgraph exploration algorithm, in which we detail how to start from the topic entity, leverage LLMs to discover relevant entities and relations in each round of subgraph exploration, and ultimately complete the entire subgraph exploration process.

\begin{algorithm}[htb]
\caption{Subgraph Exploration}\label{alg:update_Subgraph}
\begin{algorithmic}[1]
\REQUIRE $entityDict$, $entityName$, $question$, \\ $maxWidth$, $depth$, $path$

\STATE Set $originalPath$ as $path$

\IF{$depth = 0$}
    \STATE Initialize $path$ as $[ \;]*maxWidth$
\ENDIF

\FOR{$eid$ in $entityDict$}
    \STATE Find $relevantRelations$
    \FOR{$relation$ in $relevantRelations$}
        \STATE Find entities linked by $relation$
    \ENDFOR
\ENDFOR

\STATE \textbf{Extract} $relevantEntities$ using candidate entities
\STATE \textbf{Update} $path$ and $entityDict$ based on relevance

\STATE $extraPath \leftarrow (path - originalPath)$

\RETURN $extraPath$, $entityDict$

\end{algorithmic}
\end{algorithm}
\section{Path Update Algorithm}\label{appendix-Path-update}
Here we present the updating algorithm for reasoning paths, where we implement directional updates of paths. The paths generated by this algorithm can be directly visualized, which significantly enhances the interpretability of LLM question-answering results by providing observable visual subgraphs for the outcomes. 

\begin{algorithm}[htb]
    \caption{Update Reasoning Path in Subgraph}\label{alg:update_path}
    
    \begin{algorithmic}[1]
    \REQUIRE $path$, $pathIsHead$, $isHead$, $r$, $e$
    
    \IF{\textbf{not} $pathIsHead$} 
        \IF{\textbf{not} $isHead$}
            \STATE $newPath \gets path + [\leftarrow, r, \leftarrow, e]$.
        \ELSE
            \STATE $newPath \gets path + [\rightarrow, r, \rightarrow, e]$.
        \ENDIF
    \ELSE
        \IF{\textbf{not} $isHead$}
            \STATE $newPath \gets [e, \rightarrow, r, \rightarrow] + path$.
        \ELSE
            \STATE $newPath \gets [e, \leftarrow, r, \leftarrow] + path$.
        \ENDIF
    \ENDIF
    
    \STATE Append $newPath$ to $path$
    \RETURN $path$
    
    \end{algorithmic}
\end{algorithm}

\section{Efficiency Analysis}\label{Efficiency}

We present detailed experimental results demonstrating KnowPath's efficiency in Table \ref{table:tokens-WebQSP}, Table \ref{table:tokens-SimpleQuestions}, and Table  \ref{table:tokens-WebQuestions}.
\begin{table}[hbt]
\centering
\resizebox{\columnwidth}{!}{  % 
\begin{tabular}{lccccc}
\toprule
\textbf{Method} & \textbf{LLM Call} & \textbf{Total Token} & \textbf{Input Token} &\textbf{Time(s)}\\
\midrule
ToG & 15.9 & 7018.9 & 6031.2 & 63.1\\
PoG & 9 & 5517.7 & 5234.8 & 16.8\\
KnowPath & \textbf{5.59} & \textbf{2477.9} & \textbf{2223.8} & \textbf{11.2} \\
\bottomrule
\end{tabular}
}
\caption{Cost-effectiveness analysis on the WebQSP dataset between our KnowPath and the strongly prompt-driven knowledge-enhanced benchmarks (ToG and PoG).}
\label{table:tokens-WebQSP}
\end{table}

\begin{table}[hbt]
\centering
\resizebox{\columnwidth}{!}{  % 
\begin{tabular}{lccccc}
\toprule
\textbf{Method} & \textbf{LLM Call} & \textbf{Total Token} & \textbf{Input Token} &\textbf{Time(s)}\\
\midrule
ToG & 7.9 & 4666.16 & 4020.46 & 30.1\\
PoG & \textbf{4.85} & 2792.79 & 2585.19 & \textbf{7.64}\\
KnowPath & 5.35 & \textbf{1272.3} & \textbf{1064.44} & 8.84 \\
\bottomrule
\end{tabular}
}
\caption{Cost-effectiveness analysis on the SimpleQuestions dataset between our KnowPath and the strongly prompt-driven knowledge-enhanced benchmarks (ToG and PoG).}
\label{table:tokens-SimpleQuestions}
\end{table}

\begin{table}[hbt]
\centering
\resizebox{\columnwidth}{!}{  % 
\begin{tabular}{lccccc}
\toprule
\textbf{Method} & \textbf{LLM Call} & \textbf{Total Token} & \textbf{Input Token} &\textbf{Time(s)}\\
\midrule
ToG & 14.7 & 8297.35 & 7021.45 & 59.4\\
PoG & 8.1 & 6114.55 & 5788.83 & 15.6\\
KnowPath & \textbf{5.16} & \textbf{2426.1} & \textbf{2183.5} & \textbf{10.7} \\
\bottomrule
\end{tabular}
}
\caption{Cost-effectiveness analysis on the WebQuestions dataset between our KnowPath and the strongly prompt-driven knowledge-enhanced benchmarks (ToG and PoG).}
\label{table:tokens-WebQuestions}
\end{table}

\section{Datasets}\label{dataset}

We provide comprehensive statistics for the four knowledge graph QA datasets: ComplexWebQuestions~\cite{cwq} (CWQ), WebQuestionsSP~\cite{webqsp} (WebQSP), SimpleQuestions~\cite{simpleqa}, and WebQuestions~\cite{webquestion}, as presented in Table \ref{table:datasets}.
\begin{table}[htb]
\centering
\resizebox{\columnwidth}{!}{
\begin{tabular}{lcccc}
\toprule
\textbf{Dataset} & \textbf{Train} & \textbf{Test} & \textbf{Knowledge base} \\
\midrule
CWQ & 27734 & 3531 & FreeBase \\
WebQSP & 3098 & 1639 & FreeBase \\
SimpleQuestions & 75910 & 21687 & FreeBase\\
WebQuestions & 3778 & 2032 & FreeBase\\
\bottomrule
\end{tabular}
}
\caption{Statistical Information of Four KGQA Datasets.}
\label{table:datasets}
\end{table}

\section{Knowledge Subgraph Visualization}\label{appendix-Visualize}

\begin{figure}[t]
  \centering
    \includegraphics[width=0.98\linewidth]{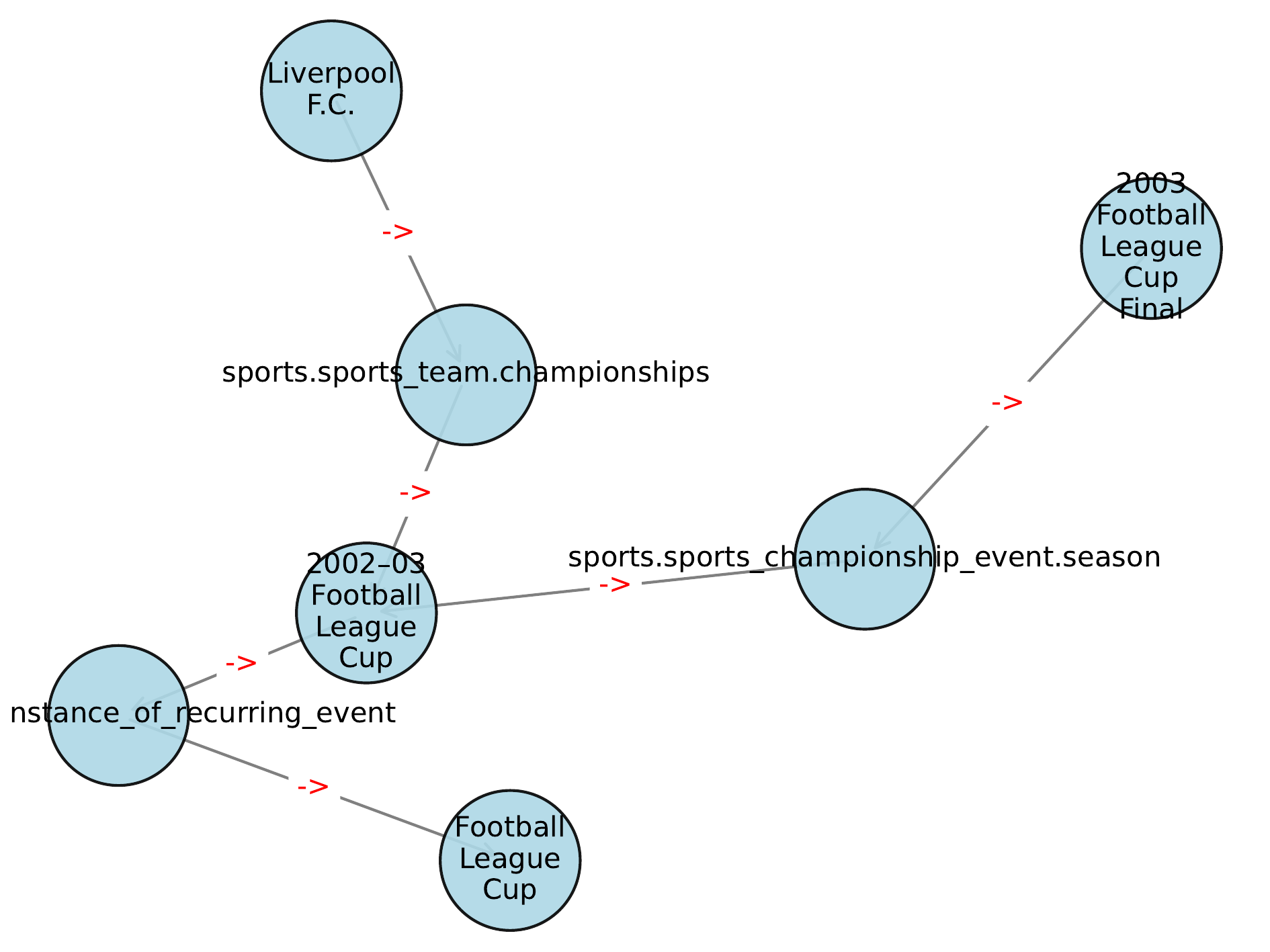}
  % \hspace{0.01\linewidth}
  \caption{Visualization results of the knowledge subgraph for Case 1 in the paper.}
  \label{fig:appendix-case1}
\end{figure}

\begin{figure}[t]
  \centering
    \includegraphics[width=0.98\linewidth]{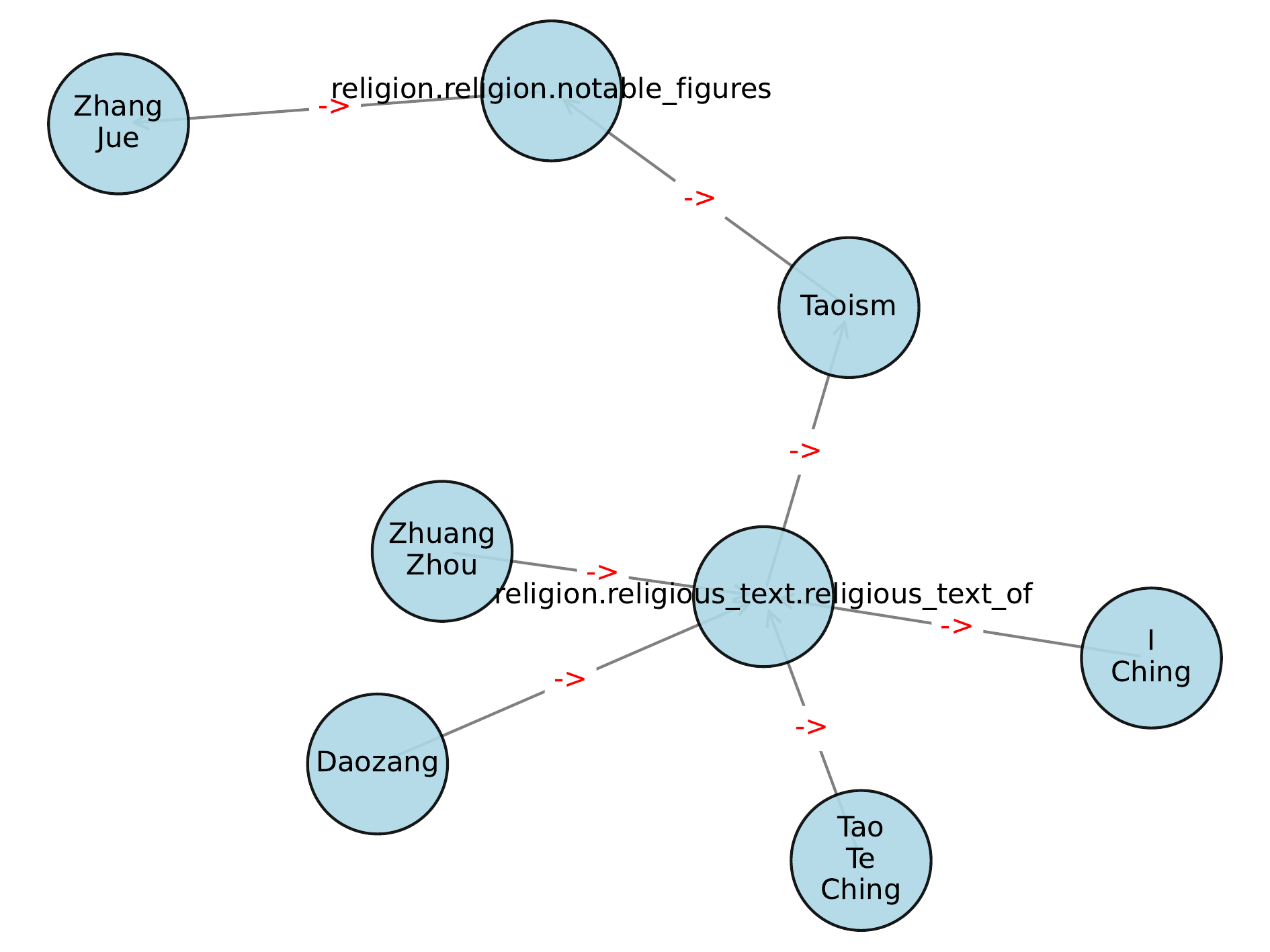}
  % \hspace{0.01\linewidth}
  \caption{Visualization results of the knowledge subgraph for Case 2 in the paper.}
  \label{fig:appendix-case2}
\end{figure}

KnowPath not only enables large language models to answer questions more accurately, but also provides interpretable knowledge subgraphs to support the answers. Thanks to our proposed Algorithm \ref{alg:update_Subgraph} and Algorithm \ref{alg:update_path}, KnowPath can directly generate the final knowledge subgraphs, significantly improving result reliability and helping mitigate hallucinations in LLMs. In Figure \ref{fig:appendix-case1} and Figure \ref{fig:appendix-case2} we present visualized subgraph examples from two cases in the main text - note these subgraphs are directly generated by code using data collected through KnowPath.


\section{Prompts}\label{appendix-prompts}

\subsection{Inference Paths Generation Examples}

\begin{appendixbox}[title={Inference Paths Generation Query}] 
    Query:what is the majority party now in the country where Canberra is located?
\end{appendixbox}

\begin{appendixbox}[title={Inference Paths Generation Triples}] 
    Here are 15 related triples based on the entities and question you provided, ranked from high to low relevance: \\
        (Country associated with Canberra, has capital, Canberra) \\
        (Country associated with Canberra, located in continent, Australia) \\
        (Country associated with Canberra, has government type, Federal parliamentary constitutional monarchy) \\
        (Country associated with Canberra, has official language, English) \\
        (Country associated with Canberra, has majority party, Australian Labor Party) \\
        (Australian Labor Party, is led by, Anthony Albanese) \\
        (Australian Labor Party, holds seats in, House of Representatives) \\
        (Australian Labor Party, holds seats in, Senate) \\
        (Australian Labor Party, has political alignment, center-left) \\
        ......
\end{appendixbox}

\begin{appendixbox}[title={Inference Paths Generation Answer}] 
    Based on the relevant triples and the query, the final answer is:\\
    Australian Labor Party\\
    The overall reasoning path, starting from the topic entity, is:\\
    Country associated with Canberra (topic entity)\\
    $\rightarrow$ has capital $\rightarrow$ Canberra\\
    $\rightarrow$ has majority party $\rightarrow$ Australian Labor Party (final answer)\\
\end{appendixbox}

\subsection{Inference Paths Generation Prompt}\label{prompt-inference}

\begin{appendixbox}[title={Inference Paths Generation Prompt}] 
    You need to answer Question using follow steps: \\
        step1: You need to extract the most relevant topic entities from the Question. \\
        step2: Based on the topic entities and Question. List the 15 related knowledge triples from high to low in terms of relevance to the Question. The triples are given in the form of (entity, relation, entity). \\
        step3: Based on the knowledge triples you listed, combined with the Question and topic entities, you need to give the final answer. In addition, you need to give the reasoning path. The overall format should be "entity1$\rightarrow$relation1$\rightarrow$entity2$\rightarrow$relation2$\rightarrow \\
        $entity3$\rightarrow$...$\rightarrow$end". \\
        The answer format is \{reasoning\_path : ["entity1$\rightarrow$relation1$\rightarrow$entity2$\rightarrow$relation2$ \\ \rightarrow$entity3$\rightarrow$...$\rightarrow$end"], "response": "based on the knowledge, the answer to the question \$question is xxxx" \}
\end{appendixbox}

\subsection{Relation Exploration Prompt}\label{prompt-Relation}
\begin{appendixbox}[title={Relation Exploration Prompt}] 
    Dict : \{ \\
        "Question" : \$question, \\
        "Topic entity" : \$topicEntity, \\
        "Knowledge Path" : \$knowpath\_str, \\
        \} \\
        RelationList: \$relationList \\
        Now you need to find out up to 7 most relevant relations from RelationList to each entry in the dictionary Dict and put them into a list called Relations. The answer format is: \{ "Relations":[xxx, xxx, xxx,...] (length up to 5) \}. Do not output any extra content except what is required by the format. \\
        Answer:
\end{appendixbox}

\subsection{Entity Exploration Prompt}\label{prompt-Entity}
\begin{appendixbox}[title={Entity Exploration Prompt}] 
    Dict : \{ \\
    "Question" : \$question, \\
    "Topic entity" : \$topicEntity, \\
    "Knowledge Path" : \$knowpath\_str, \\
    "RelationList" : \$relationList, \\
    \} \\
    EntityList: \$entityList \\
    Now you need to find out up to 7 entities that are most relevant to each entry in the dictionary Dict from EntityList by relevance, and put them into a list called Entities. The answer format is: \{ "Entities":[xxx, xxx, xxx,...] (length up to 5) \}. Do not output any extra content except what is required by the format. \\
    Answer:
\end{appendixbox}

\subsection{Evaluation-based Answering Prompt}\label{prompt-Evaluation}
\begin{appendixbox}[title={Evaluation-based Answering Prompt}] 
    Reasoning\_path:\$subgraph \\
        Based on the Reasoning\_path and your own knowledge, you need to determine whether the Question:\$question can be answered. '->' and '<-' indicate the direction of Reasoning\_path between entities and relationships. \\
        Requests: \\
        1. The answer format is: \{ "Answerable": True or False, "Response": "the answer to the question \$question is xxxx" \} \\
        Answer:
\end{appendixbox}

\begin{appendixbox}[title={CoT Prompt}] 
    Q: What state is home to the university that is represented in sports by George Washington Colonials men's basketball? \\
        A: First, the education institution has a sports team named George Washington Colonials men's basketball in is George Washington University, Second, George Washington University is in Washington D.C. \\
        The answer is {Washington, D.C.}.
        \\
        \\
        Q: Who lists Pramatha Chaudhuri as an influence and wrote Jana Gana Mana? \\
        A: First, Bharoto Bhagyo Bidhata wrote Jana Gana Mana. Second, Bharoto Bhagyo Bidhata lists Pramatha Chaudhuri as an influence. \\
        The answer is {Bharoto Bhagyo Bidhata}.
        \\
        \\
        Q: Who was the artist nominated for an award for You Drive Me Crazy? \\
        A: First, the artist nominated for an award for You Drive Me Crazy is Britney Spears. \\
        The answer is {Jason Allen Alexander}.
        \\
        \\
        Q: What person born in Siegen influenced the work of Vincent Van Gogh? \\
        A: First, Peter Paul Rubens, Claude Monet and etc. influenced the work of Vincent Van Gogh. Second, Peter Paul Rubens born in Siegen. \\
        The answer is {Peter Paul Rubens}.
        \\
        \\
        Q: What is the country close to Russia where Mikheil Saakashvii holds a government position? \\
        A: First, China, Norway, Finland, Estonia and Georgia is close to Russia. Second, Mikheil Saakashvii holds a government position at Georgia. \\
        The answer is {Georgia}.
        \\
        \\
        Q: What drug did the actor who portrayed the character Urethane Wheels Guy overdosed on? \\
        A: First, Mitchell Lee Hedberg portrayed character Urethane Wheels Guy. Second, Mitchell Lee Hedberg overdose Heroin. \\
        The answer is {Heroin}.
        \\
        \\
        Q: \$question \\
        A: \\
\end{appendixbox}

\section{SPARQL Queries}\label{Sparql}
We employ SPARQL queries to retrieve entities and relations from the Freebase knowledge graph. The complete set of SPARQL statements utilized by KnowPath is provided below, with implementation details available in our publicly released codebase.

\begin{appendixbox}[title={Head Relation Search}] 
    PREFIX ns: <http://rdf.freebase.com/ns/> \\
        SELECT DISTINCT ?relation \\
        WHERE \{ \\
            ns:\%s ?relation ?tail . \\
        \} \\
\end{appendixbox}

\begin{appendixbox}[title={Tail Relation Search}] 
    PREFIX ns: <http://rdf.freebase.com/ns/> \\
        SELECT DISTINCT ?relation \\
        WHERE \{ \\
            ?head ?relation ns:\%s . \\
        \} \\
\end{appendixbox}

\begin{appendixbox}[title={Head Entity Search}] 
    PREFIX ns: <http://rdf.freebase.com/ns/> \\
        SELECT DISTINCT ?Entity \\
        WHERE \{ \\
            ns:\%s ns:\%s ?Entity . \\
        \} \\
\end{appendixbox}

\begin{appendixbox}[title={Tail Entity Search}] 
    PREFIX ns: <http://rdf.freebase.com/ns/> \\
        SELECT DISTINCT ?Entity \\
        WHERE \{ \\
            ?Entity ns:\%s ns:\%s . \\
        \} \\
\end{appendixbox}

\end{document}